# Site-Level Fine-Tuning with Progressive Layer Freezing: Towards Robust Prediction of Bronchopulmonary Dysplasia from Day-1 Chest Radiographs in Extremely Preterm Infants

Goedicke-Fritz, Sybelle[1#]; Bous, Michelle[1#]; Engel, Annika[2#]; Flotho, Matthias[2,5]; Hirsch, Pascal[2]; Wittig, Hannah[1]; Milanovic, Dino[2]; Mohr, Dominik[1]; Kaspar, Mathias[6]; Nemat, Sogand[3]; Kerner, Dorothea[3]; Bücker, Arno [3]; Keller, Andreas[2,5,7]; Meyer, Sascha[4]; Zemlin, Michael[1+]; Flotho, Philipp[2,5+*]

[#] These authors contributed equally to this work, share the first authorship and can cite this paper with their name first.

[+] The last authors contributed equally to this work, share the senior authorship and can cite this paper with their name last.

[*] Corresponding author

[1] Department of General Paediatrics and Neonatology, Saarland University, Campus Homburg, Homburg/Saar, Germany

[2] Chair for Clinical Bioinformatics, Saarland Informatics Campus, Saarland University, 66123 Saarbrücken, Germany

[3] Department of Radiology and Interventional Radiology, University Hospital of Saarland, Homburg, Germany.

[4] Clinical Centre Karlsruhe, Franz-Lust Clinic for Paediatrics, Karlsruhe, Germany

[5] Helmholtz Institute for Pharmaceutical Research Saarland (HIPS), Saarland University Campus, Germany

[6] Digital Medicine, University Hospital of Augsburg, Augsburg, Germany

[7] Pharma Science Hub (PSH), Saarland University Campus, Germany

*Corresponding author:* philipp.flotho@uni-saarland.de (P. Flotho)

## Highlights

[1] Day-1 chest radiograph predicts later lung disease risk (area under curve ≈ 0.78).

[2] Chest-x-ray pretraining beats natural-image pretraining for neonatal scans.

[3] Progressive layer freezing with a brief linear probe works best.

[4] Routine respiratory distress grades only weakly predict later lung disease.

**Abstract**

Bronchopulmonary dysplasia (BPD) is a chronic lung disease affecting 35% of extremely low birth weight infants. Defined by oxygen dependence at 36 weeks postmenstrual age, it causes lifelong respiratory complications. However, preventive interventions carry severe risks, including neurodevelopmental impairment, ventilator-induced lung injury, and systemic complications. Therefore, early BPD prognosis and prediction of BPD outcome is crucial to avoid unnecessary toxicity in low risk infants.

Admission radiographs of extremely preterm infants are routinely acquired within 24h of life and could serve as a non-invasive prognostic tool. In this work, we developed and investigated a deep learning approach using chest X-rays from 163 extremely low-birth-weight infants (≤32 weeks gestation, 401-999g) obtained within 24 hours of birth. We fine-tuned a ResNet-50 pretrained specifically on adult chest radiographs, employing progressive layer freezing with discriminative learning rates to prevent overfitting and evaluated a CutMix augmentation and linear probing.

For moderate/severe BPD outcome prediction, our best performing model with progressive freezing, linear probing and CutMix achieved an AUROC of 0.78 ± 0.10, balanced accuracy of 0.69 ± 0.10, and an F1-score of 0.67 ± 0.11. In-domain pre-training significantly outperformed ImageNet initialization ($p = 0.031$) which confirms domain-specific pretraining to be important for BPD outcome prediction. Routine IRDS grades showed limited prognostic value (AUROC 0.57 ± 0.11), confirming the need of learned markers.

Our approach demonstrates that domain-specific pretraining enables accurate BPD prediction from routine day-1 radiographs. Through progressive freezing and linear probing, the method remains computationally feasible for site-level implementation and future federated learning deployments.

## Introduction

Bronchopulmonary dysplasia (BPD) is a severe chronic lung disease affecting approximately 35% of extremely low birth weight infants (ELBW, <1,000g) (Gortner et al., 2011; Smith et al., 2005; Stichtenoth et al., 2012; Thomas & Speer, 2005). Defined as the need for oxygen supplementation at 36 weeks postmenstrual age (Herting, 2013; Jobe & Bancalari, 2001), BPD compromises the respiratory system and is associated with severe long-term sequelae and high morbidity (Jobe & Bancalari, 2001; Schmidt et al., 2003). The current diagnostic framework, based on retrospective oxygen dependency assessment, provides no early prognostic capability when interventions would be most effective. The pathogenesis of BPD is considered multifactorial. Immaturity of the preterm infants' lungs causes alterations in anatomy and biochemistry of the lungs (Baraldi & Filippone, 2007). The current classification of the severity of BPD is based on the recommendations of Jobe et al. depending on whether there is still need for oxygen at a postmenstrual age of 36 weeks (Herting, 2013; Jobe & Bancalari, 2001).

Prematurity in general affects approximately 10% of all children and results in drastically altered antigen exposure due to premature confrontation with microbes, nutritional antigens, and other environmental factors. During the last trimester of pregnancy, the fetal immune system undergoes critical adaptations to tolerate maternal and self-antigens, while simultaneously preparing for postnatal immune defense through the acquisition of passive immunity via maternal antibodies. Since the perinatal period is regarded as the most important "window of opportunity" for immune and metabolic imprinting, preterm birth may significantly impact the development of immune-mediated diseases later in life (Goedicke-Fritz et al., 2017). Diagnostics of various risks are often difficult in premature infants; blood samples, in particular, can only be taken to a limited extent as painful procedures such as venipuncture increase the risk of cerebral hemorrhage and anemia requiring transfusion in very small premature infants (Zea-Vera & Ochoa, 2015; Zemlin et al., 2019) also often take a long time (e.g. the determination of interleukin-8 takes at least 70 minutes). Classic neonatal conditions include intraventricular hemorrhage (IVH), retinopathy prematurorum (ROP), periventricular leukomalacia (PVL) and necrotizing enterocolitis (NEC), all of which are associated with potential long-term consequences or are even fatal for premature babies.

Given the multifactorial pathogenesis and variable clinical trajectory of BPD, decisions regarding early therapeutic interventions must be made with caution. Misclassification of an infant's true risk for BPD may lead to unnecessary treatment in neonates who would not have developed the disease. Surfactant administration, particularly when performed via endotracheal intubation, carries risks such as bradycardia, hypoxia, laryngospasm, and pneumothorax – especially in infants without significant surfactant deficiency who may not benefit clinically from the treatmen (Polin et al., 2014; Sweet et al., 2023). Additionally, improved lung compliance following surfactant instillation can sometimes result in lung overdistension and unstable ventilation parameters (Sardesai et al., 2017).

Similarly, the early use of systemic corticosteroids (e.g., dexamethasone or hydrocortisone), although beneficial in selected high-risk infants, must be carefully weighed (Doyle, 2021). Studies have shown an association between postnatal steroid use and adverse neurodevelopmental outcomes, including increased risk of cerebral palsy and impaired brain growth (Doyle et al., 2014; Yeh et al., 2004). Short-term side effects such as hyperglycemia, hypertension, gastrointestinal perforation, and increased infection risk are also well documented (Watterberg et al., 2004). Recent findings from a large European cohort study further emphasize the importance of risk stratification when considering postnatal steroid (PNS) treatment to prevent BPD. The study demonstrated that PNS use was associated with an increased risk of gross motor impairment at two years of corrected age in infants with a high BPD risk (OR 1.95, 95% CI 1.18–3.24; $p = 0.010$), regardless of the type of steroid used. Notably, cognitive anomalies were particularly increased in infants treated with dexamethasone or betamethasone, whereas this association was not observed with hydrocortisone (Nuytten et al., 2020).

Moreover, unnecessary mechanical ventilation poses an independent risk factor for iatrogenic BPD. Ventilator-induced lung injury via volutrauma, barotrauma, and atelectrauma, combined with oxygen toxicity, can trigger inflammation and oxidative damage to the immature lung (Dini et al., 2024; Slutsky, 1999). Hyperoxia itself is associated with long-term complications such as retinopathy of prematurity and increased pulmonary reactivity (Askie et al., 2003).

Currently, there is no reliable method for early detection of BPD, and novel prognostic tools for BPD are of great medical interest, because early identification of at-risk infants could allow for timely and individualized preventive interventions without exposing healthy children to the toxicity of such interventions. BPD remains a major cause of morbidity and mortality in extremely preterm infants, and its multifactorial pathogenesis involves antenatal, perinatal, and postnatal risk factors such as intrauterine growth restriction, inflammation, mechanical ventilation, and oxygen toxicity (Higgins et al., 2018; Jensen & Schmidt, 2014; Jobe & Bancalari, 2001). Since many of the available preventive strategies – such as less invasive surfactant administration, early caffeine therapy, and lung-protective ventilation—are most effective when applied within the first hours or days of life (Polin et al., 2014; Schmidt et al., 2006; Sweet et al., 2023), the ability to predict BPD shortly after birth is of critical clinical relevance.

In this context, artificial intelligence (AI)-based analysis has emerged as a promising approach to prognose BPD development from clinical parameters and radiographic images (Ali et al., 2025; Cho et al., 2025; Chou et al., 2024; Kwok et al., 2023; Xing et al., 2022). Such tools have the potential to complement clinical judgment, reduce variability in care, and guide evidence-based decision-making during a highly sensitive therapeutic window. However, most approaches for the assessment of BPD risk rely exclusively on clinical variables and controlled evaluation of imaging-only deep-learning pipelines started gaining attention in the past years (Ali et al., 2025; Cho et al., 2025; Chou et al., 2024; Hwang et al., 2023; Kwok et al., 2023; Xing et al., 2022).

Deep learning has transformed medical imaging, with CNNs achieving expert-level performance across diagnostic tasks. In the context of BPD, Xing et al. trained on day-28 chest radiographs and demonstrated that BPD severity is learnable (Xing et al., 2022). Chou et al. (Chou et al., 2024) include radiographs recorded only 24 hours after birth with a pipeline containing lung mask segmentation and Ali et al. (Ali et al., 2025) benchmarked thirteen modern architectures on day 3, day 7, day 14 and day 28 chest-radiographs. They compared different CNN backbones initialized with RGB weights. Outside imaging, Verder et al. demonstrated that combining routine clinical variables with spectroscopic gastric-aspirate features could predict later BPD, highlighting the disease's multifactorial signature (Verder et al., 2021).

A challenge in applying deep learning on radiographic and biomedical images is the relatively limited size of datasets. Most approaches for deep learning on X-ray images therefore resorts to pretraining on RGB images and use models pretrained on large natural image corpora such as ImageNet (Russakovsky et al., 2015). Few-shot learning methods (Ahmed et al., 2024; Kukleva et al., 2024; Lin et al., 2022; Shvetsova et al., 2023) demonstrate how careful backbone adaptation, via orthogonality constraints, cross-modal conditioning or unsupervised style transfer, can stretch tiny annotation budgets, yet they still hinge on a backbone that has first been tuned on images drawn from the same visual domain.

However, combining images from different domains is not impossible: In the context of real-world RGB–thermal collections such as (Flotho et al., 2021) and the follow-up T-FAKE study (Flotho et al., 2025) that resorts to synthesising thermal images to bridge the residual modality gap – which works only because RGB and thermal photographs share natural-image statistics, a similarity that paediatric chest X-rays decisively lack.

Cohen et al. (Cohen et al., 2020) have shown that a model lifted wholesale from RGB ImageNet (Russakovsky et al., 2015) collapses on grayscale radiographs unless it is fine-tuned on in-domain X-ray data first, underscoring the need for the radiograph-specific adaptation strategy we pursue here.

In this study we analyse chest radiographs taken within the first 24 hours of life for BPD. We propose an improved training strategy with optimized feature freezing to fine tune models pretrained on large X-Ray datasets by means of TorchXRayVision (Cohen et al., 2020; Cohen et al., 2022). We demonstrate superior performance of in domain pretraining in combination with feature freezing and ablate different image augmentation approaches. Our ablation demonstrates that the performance gain stems almost entirely from two choices (see Figure 1):

i. starting from in-domain chest-radiograph weights instead of conventional ImageNet weights, and
ii. adopting a progressive layer-freezing schedule with linear probing in which only the final three residual blocks are updated, rather than training the full network.

## Material and Methods

### Patient recruitment

In this study, we analysed chest X-rays of patients recruited for the NeoVitaA trial at the Homburg/Germany site (n = 170; ethics approval number 105/14). Of these, 7 patients had to be excluded due to missing labels, meta-information or death before 36+0 weeks. The NeoVitaA trial is a prospective, multicentre, randomised, placebo-controlled, double-blind phase 3 study investigating whether high-dose oral vitamin A supplementation during the first 28 days of life in extremely preterm infants (birthweight 401–999 g; gestational age ≤32+0 weeks) reduces the incidence of bronchopulmonary dysplasia (BPD) or death compared to standard care (Meyer et al., 2024; Meyer & Gortner, 2017). Recruitment for the NeoVitaA trial was completed in 2022.

The retrospective analysis of the chest X-ray images from this trial was conducted in accordance with the Declaration of Helsinki and received separate ethical approval under reference number 72/14. Thus, while the original trial was approved under ethics reference 105/14, this subsequent retrospective image analysis was independently approved under ethics reference 72/14.

Infants were randomised to receive either 5000 IU/kg/day of enteral vitamin A (Vitadral® oral drops) or placebo for 28 days, in addition to routine vitamin A supplementation. Vitamin A supplementation in preterm infants plays a crucial part in lung growth and differentiation; intramuscular vitamin A application has been shown to decrease BPD rates (Darlow et al., 2016; Tyson et al., 1999). Nevertheless, intramuscular application is painful and therefore obsolete in clinical practice. However, the effect of postnatal additional high-dose oral vitamin A in extremely low birth weight (ELBW) infants supplemented within the first 28 days of life is investigated by the NeoVitaA study cohort. The study found no significant difference in the rate of moderate or severe bronchopulmonary dysplasia or death between the high-dose vitamin A group and the control group, and serum retinol levels remained similar in both groups (Meyer et al., 2024; Meyer & Gortner, 2017). Despite decades of neonatal research, multiple interventions – including permissive hypercapnia, inhaled nitric oxide, and high-dose vitamin A – have repeatedly failed to lower the incidence of bronchopulmonary dysplasia in very preterm infants (Barrington et al., 2017; Ma & Ye, 2016; Meyer et al., 2024; Meyer & Gortner, 2017; Meyer et al., 2014).

Inclusion for the present analysis required the availability of a chest radiograph obtained within the first 24 hours of life as part of initial respiratory assessment. Infants were excluded if they had major congenital anomalies (e.g., congenital heart defects, neural tube defects, gastrointestinal malformations), signs of non-bacterial infection at birth, or if no adequate imaging or clinical data were available. All radiographs had been acquired as part of standard clinical care upon admission to the neonatal intensive care unit (Meyer et al., 2024; Meyer & Gortner, 2017).

## Diagnosis of Bronchopulmonary Dysplasia (BPD)

In this study, the diagnosis of bronchopulmonary dysplasia (BPD) was based on a modified version of the National Institute of Child Health and Human Development (NICHD) definition proposed by Jobe and Bancalari (Jobe & Bancalari, 2001). BPD was diagnosed in infants who required supplemental oxygen or respiratory support for at least 28 days and were then assessed at 36+0 weeks postmenstrual age (PMA) to determine disease severity.

BPD can be classified into mild, moderate and severe (Herting, 2013; Jobe & Bancalari, 2001) and infants were categorized as follows:

- **Mild BPD**: Need for respiratory support for ≥28 days, but breathing room air at 36+0 weeks PMA.
- **Moderate BPD**: Need for respiratory support for ≥28 days and receiving supplemental oxygen with a fraction of inspired oxygen ($FiO_2$) >0.21 but <0.30 at 36+0 weeks PMA.
- **Severe BPD**: Need for respiratory support for ≥28 days and either $FiO_2$ ≥0.30 or ongoing positive pressure respiratory support (e.g., CPAP or mechanical ventilation) at 36+0 weeks PMA.

This standardized definition enabled a consistent and clinically relevant assessment of pulmonary outcomes in extremely preterm infants. Infants who died before 36+0 weeks PMA were excluded from the study and infants who died after the 36+0 weeks were classified with the respective BPD outcome based on respiratory support.

## Assessment of IRDS

To assess the presence of IRDS, chest radiographs taken during initial respiratory stabilization within the first 24 hours after birth were analyzed. The images were independently reviewed by two board-certified pediatric radiologists and one neonatologist, all blinded to clinical outcomes. Any discrepancies in interpretation were resolved by consensus discussion. To create binary labels for deep learning model training, we computed the majority vote among the three experts and then binarized the consensus: grades I-II (stages 1-2) were labeled as mild/no IRDS, while grades III-IV (stages 3-4) were labeled as moderate/severe IRDS.

The diagnosis of IRDS was based on predefined radiographic criteria characteristic of the condition, such as a fine reticulogranular pattern, prominent air bronchograms, and reduced lung volumes, as described in the established literature. The corresponding X-ray images were already available, as they have been taken as a part of routine care at admission on neonatal intensive care unit (NICU).

IRDS severity is graded in four stages, where each stage is characterized by unique visual features and the stage guides surfactant application. Stage I is characterized by fine, evenly distributed, reticulo-granular haze which results from air in the bronchioles, stage II has the same granular pattern plus streaky/patchy opacities and prominent air-bronchograms that now extend beyond the heart shadow. In stage III opacity increases together with the presence of granular textures and heart and diaphragmatic borders become indistinct. Stage IV characterizes near-complete opacification together with very low lung volumes (Prodanovic et al., 2024).

**Data Splitting and Preprocessing**

We obtained chest X-ray images from our local NICU for ELBW infants. The images where manually cropped and aligned to isolate the upper chest region. Preprocessing was performed in accordance with the two main pretraining strategies: namely different normalization for RGB and X-ray pretraining.

Due to the very low negative cases and overall case count, we binned mild BPD together with the BPD negative cases and moderate BPD together with the severe BPD cases which also maps required medical interventions.

Our primary evaluation of model generalization was conducted using repeated 5-fold cross-validation where the patient cohort was split into five balanced folds and this process was repeated six times with different random seeds. The folds were drawn on the smaller, positive cohort and filled uniquely with randomly drawn negative samples to create balanced test sets. Splits were created on patient level and only a single image per patient was included in the test sets while all available images were used during training. Particularly, our splitting strategy ensured that all BPD positive patients were used exactly once during testing, no BPD negative patient was used more than once during testing and no images of the same patient was present in both test and train set for any given split.

### Machine Learning Approach

We employed a transfer learning strategy based on ResNet-50 (He et al., 2016) processing the images directly without prior lung segmentation. For our primary models, we used pretrained weights from large-scale chest X-ray datasets (ChestX-ray8 (Wang et al., 2017), PadChest (Bustos et al., 2020), MIMIC-CXR (Johnson et al., 2019), RSNA Pneumonia Detection Challenge dataset (Shih et al., 2019), SIIM–ACR Pneumothorax Segmentation (Zawacki et al., 2019), Indiana-University / Open-I Collection (Demner-Fushman et al., 2016), VinDr-CXR (Nguyen et al., 2022)). Generally, the early layers learn generic edge and texture filters, while deeper layers capture domain-specific features (e.g., lung fields, heart silhouette, pathological patterns).

The model architecture was adapted for binary classification by replacing the final fully-connected layer with a new single-unit output head. Our fine-tuning process utilized two main strategies.

Some experiments began with an initial linear probing phase, where only the newly added classifier head was trained for several epochs while the entire pretrained backbone remained frozen. This was followed by a progressive unfreezing schedule inspired by ULMFiT (Joseph & Joshi, 2024), where we sequentially unfroze deeper layers of the network (layer 4, followed by layer 3, then layer 2) with discriminative learning rates. This two-stage approach allows the model to first learn the classification task with stable, powerful features before adapting those features to our specific dataset, minimizing the risk of catastrophic forgetting. Furthermore, we include 10 epochs of linear probing before fine tuning (Ke et al., 2021; Kornblith et al., 2019).

Training was performed using the AdamW optimizer with a Binary Cross-Entropy with Logits loss function. To manage class imbalance in the training set, we used weighted random sampling which ensures that each batch contained a balanced amount of samples from both classes. We utilized a OneCycle learning rate schedule, which gradually increases the learning rate for the first part of training before annealing it, to facilitate faster convergence. All training was conducted using mixed-precision to improve computational efficiency.

Our augmentation pipeline included random rotations, flips, partial random masking of pixels, and a random Gaussian blur to improve model robustness. Furthermore, we ablated the usefulness of a CutMix augmentation (Yun et al., 2019) as well as ImageNet (Russakovsky et al., 2015) (RGB) pretraining that we found as part of related pipelines for BPD prediction (see Figure 1 for an overview of our approach).

## Results

We trained our model for 30 epochs without and 30 / 40 epochs with CutMix augmentation and ablated the chosen freezing schedule for fine-tuning. Our core insight is that Res-Net pretrained on RGB images performs significantly worse than the models pretrained on the X-ray datasets.

Our proposed progressive freezing schedule, when applied to the X-ray pretrained backbone together with linear probing, achieved the highest overall discriminative power, averaging a mean AUROC of 0.783 ± 0.095 and a mean balanced accuracy of 0.686 ± 0.101. The overall highest accuracy is achieved with progressive layer freezing without linear probing.

### Impact of Pre-training and Fine-tuning Strategies

The possibility of BPD prediction from radiographs has been proposed before, however, there is a lack of ablations to guide model choice, pre-training, and fine-tuning strategies on the often-small datasets available. Here, we investigate whether pre-training on out-of-domain data (ImageNet RGB images) versus in-domain data (adult chest radiographs) boosts model performance and whether different fine-tuning strategies improve model stability.

Our results show that the choice of pre-training domain is the most critical factor (see Figure 2). Models pretrained on in-domain chest X-ray data consistently and significantly outperformed their counterparts that started from ImageNet weights.

The baseline XRV-ProgFreeze (ResNet-50 with layers ≤3 frozen) achieved a mean AUROC = 0.775 ± 0.097, whereas its ImageNet-initialised twin (RGB-ProgFreeze) reached only 0.717 ± 0.094 (paired t, six outer-repeat means, $p = 0.0309$). Focusing on the different fine-tuning strategies for X-ray pretraining, the differences become roughly five-fold smaller (see Table 1).

However, adding a brief linear-probing warm-up, progressive layer unfreezing, and CutMix augmentation increased the mean AUROC to 0.783 ± 0.095 (XRV-ProgFreeze + LP + CutMix, 40 epochs). This configuration was significantly better than the same backbone tuned with linear probing alone and also outperformed a fully unfrozen counterpart trained with identical augmentation stack ($p < 0.05$). Furthermore, using the CutMix augmentation benefited significantly from increasing training time from 30 to 40 epochs ($p = 0.0195$).

In practical terms, securing an in-domain backbone is the prerequisite for reliable BPD prediction on small neonatal datasets; once this is in place, the choice among the tested fine-tuning schedules should be guided by computational budget and ease of deployment rather than by expectations of large additional accuracy gains.

This suggests that as long as an appropriate in-domain pretrained model is used, several fine-tuning approaches can achieve comparable top-tier performance, and the specific freezing schedule or fine-tuning strategy is less critical than the initial weight selection.

**BPD prediction from IRDS**

To analyse the joint information content in IRDS and BPD, we built a simple prognostic model that tries to predict later BPD outcome from the three ordinal radiographic scores that neonatologists record in the first 24 h for neonatal RDS. Of 170 infants with a recorded BPD outcome, 9 lacked one or more IRDS ratings and were excluded; 161 remain (57 moderate/severe, 104 none/mild).

Each of the 161 infants contributes one row with the three expert grades and the final BPD outcome (57 moderate/severe, 104 none/mild). We evaluated the model with five balanced folds, repeated five times in the same way as before: We split the 57 positive patients into five disjoint subsets with 11–12 infants each. For every subset we draw an equal-sized, non-overlapping sample of negative patients and use that pair as the test set for one fold. This way, each positive sample (smaller class) is used once for testing while ensuring balanced test splits. The classifier is a degree-two polynomial logistic regression with an L2 penalty and class-balanced weights to model simple interactions between the three scores while keeping the parameter count (nine) well below the effective sample size.

Across the 25 balanced folds the area under the ROC curve is 0.573 ± 0.105 and the balanced accuracy is 0.564 ± 0.099; the 95 % confidence intervals of both metrics stay just above 0.50. This demonstrates that the IRDS grades carry real but weak prognostic signals, and the gain is roughly six percentage points over chance.

**IRDS prediction from X-Ray**

Next, we investigate how well the IRDS scores can be approximated with the same models from the images (see Figure 3 and Table 2). To make the outcome comparable with our previous classifier and trainable with the small dataset size, we bin the RDS scores I + II as well as III + IV.

The best model achieved a high mean test accuracy, with the top-performing configuration reaching an accuracy of 0.802 ± 0.071, a corresponding F1-score of 0.829 ± 0.076. However, while demonstrating high sensitivity (often >0.83) it has consistently low specificity (typically <0.6).

This suggests the model is highly effective at learning visual markers of severe disease (Grades III and IV) but struggles to reliably differentiate these from cases with mild or no disease (Grades I and II).

This is not necessarily a model failure but rather a reflection of the clinical ambiguity inherent in the grading scheme itself. Furthermore, binning grade I and II was motivated with practical considerations and there is no clinical reason. The visual features for severe IRDS are distinct and easily learned, whereas the boundary between mild (Grade II) and moderate (Grade III) disease is notoriously subjective. This ambiguity creates a challenging learning signal, causing the model to frequently misclassify borderline or mild cases.

This means, the model successfully learns to detect clear signs of severe pathology. The clear clinical features seem to make this easier than the prediction of severe BPD cases. At the same time, the low specificity and unstable AUROC mirrors difficulties or ambiguities in distinguishing less severe cases.

## Discussion

BPD remains one of the most serious chronic complications among extremely preterm infants. Despite numerous advances in neonatal care, early and reliable prediction of BPD is still lacking and broadly used diagnostic definitions require a retrospective confirmation of oxygen dependency at 36 weeks PMA. In this work, we have demonstrated the feasibility of BPD outcome prediction from small datasets of day-1 radiographs.

This ability to stratify infants into high- and low-risk groups based solely on a day-1 chest radiograph could facilitate more personalized care: High-risk infants may benefit from intensified monitoring or early therapeutic interventions (e.g., caffeine, non-invasive ventilation strategies), whereas low-risk infants could potentially avoid overtreatment. Admission-day chest radiographs are already a part of routine respiratory assessment in nearly all NICUs (Meyer et al., 2024) and offer a unique opportunity for early prediction of pulmonary outcomes.

Xing et al. were among the first to demonstrate the feasibility day-28 radiographs and ImageNet-initialised ResNet-50 to reach AUROC of around 0.82, confirming that overt morphologic change is predictive but leaving the question of day-1 feasibility unanswered (Xing et al., 2022). Chou et al. (Chou et al., 2024) demonstrated a very high performance on day 1 radiographs. However, inconsistencies between reported sensitivity and specificity values across their tables and unspecified data splitting limit direct comparability.

Ali et al. (Ali et al., 2025) benchmarked thirteen ImageNet-initialised CNN backbones on day-3, 7, 14 and 28 radiographs and concluded that architecture choice outweighed imaging day for early prediction. Their approach is currently state-of-the-art for the prediction of a later BPD diagnosis from different day-of-life windows of X-ray images and their best model in terms of AUROC (NASNet_Mobile, day 14) reached an AUROC of 0.8, while their ResNet-50 achieved the best performance at day 7 (AUROC of 0.672). In our cohort, we obtain an AUROC of 0.783 ± 0.095 for radiographs recorded <24h of life with a ResNet-50 that differs only in its chest-X-ray pre-training and finetuning strategy. Even the RGB/ImageNet baseline in our ablation (AUROC = 0.75) exceeds all their models trained on day 3 and we clearly outperformed their experiments with the same model. However, we cannot disentangle dataset, training and model contributions here. Motivated by the risk of long-term morbidities (Han et al., 2021; Jeon et al., 2021), our labelling (negative + mild against moderate + severe) can be classified as BPD severity prediction while Ali et al. predict presence and absence of BPD.

The recent multi-centre study by Cho et al. used 43338 neonatal CXRs to train a multi-class ResNet-50 and reported BPD-class F1 score of 0.92 on a fixed train and test split (Cho et al., 2025). However, their study does not specify post-natal age nor a day-of-life window during which the BPD images were taken and they do not explicitly exclude images taken after BPD diagnosis. Therefore, it cannot answer the early prognostic power of their model.

In the current literature, radiograph-based BPD predictors start from generic ImageNet weights and either freeze the stem or fine-tune the entire backbone in one step and we are unaware of domain-specific chest-radiograph pretraining. Our ablation therefore isolates, for the first time, the independent gains from switching to an in-domain TorchXRayVision encoder. Moreover, our findings underscore the relevance of domain-specific pretraining and progressive fine-tuning strategies for small neonatal datasets. Large-scale pretraining on chest X-rays, linear probing and parameter efficient finetuning significantly outperformed RGB pretraining.

Our results align with earlier transfer-learning research. Kornblith et al. showed that a simple linear classifier fit on frozen ImageNet features already transfers competitively across 12 vision datasets, and that a brief linear-probe stage before unfreezing improves stability of subsequent fine-tuning (Kornblith et al., 2019). Conversely, Ke et al. found that on the CheXpert (Irvin et al., 2019) benchmark ImageNet (Russakovsky et al., 2015) top-1 accuracy no longer predicts radiograph performance and that the biggest gains come from which weights are used for initialization – smaller backbones in particular profit most from pre-training (Ke et al., 2021). Future directions could include modern architectures and multimodal models for X-Ray downstream tasks such as (Chaves et al., 2024).

When comparing our day-1 X-ray model with an AUROC of 0.783 that does not rely on any clinical parameters with the original NICHD day-1 calculator as a clinical risk prediction model (Laughon et al., 2011), the derivation C-statistic for the composite outcome BPD or death was 0.793, rising to 0.854 by day 28; external validations place the day-1 value between 0.77 and 0.84, depending on cohort and outcome definition (Kanagaraj et al., 2025; Laughon et al., 2011).

A meta-analysis reports a median external C-statistic of 0.77, underscoring the transportability problem (Romijn et al., 2023). Performance of the revised 2022 NICHD estimator has been uneven: in a 223-infant US cohort it correctly classified 74 % of infants without BPD but only 6 % with Jensen Grade 2 and none with Grade 3 (Srivatsa et al., 2023), whereas a recent Canadian validation still recorded day-1 AUROC of 0.803 for death/Grade 2–3 (Kanagaraj et al., 2025). Among contemporary birth-prediction scores, the Baud model (seven clinical variables) achieved an external AUROC 0.85 (Baud et al., 2021).

While direct comparison with the NICHD BPD outcome estimator was not feasible in our cohort as we lacked the specific respiratory support modalities, that our single-image approach matches these figures argues that early lung morphology already embeds a risk signal of comparable magnitude (Laughon et al., 2011). Radiographs are routinely acquired, are less prone to documentation error, and, hence, add no data-collection burden. At the same time, image acquisition does not come with the same organizational overhead that comes with collecting and inputting multiple clinical parameters required by clinical calculators. Furthermore, our results indicate that this can already be achieved with access to relatively small cohorts at site-level.

The constrained dataset sizes typical of neonatal imaging present a fundamental challenge for deep learning deployment. However, our progressive unfreezing approach offers a natural solution for federated learning scenarios where individual sites possess insufficient data for model training. The method's selective parameter updates enable collaborative model development while preserving local data governance which is a critical requirement in pediatric healthcare.

Our approach enables a three-stage federated framework: centralized linear probing for cold-start performance, progressive unfreezing for local adaptation, and site-specific fine-tuning for scanner calibration. This addresses key distributed learning challenges such as client drift, communication efficiency: Frozen backbone features with average pooled activations would be used for classification training, which in turn would be streamed and averaged by a central server (e.g. compare (Legate et al., 2023)). This results in a global cold-start BPD classifier. In cases where labels cannot be shared server-side, linear probing could be performed locally and subsequently aggregated with FedAvg (McMahan et al., 2017). However, this approach would potentially face convergence challenges (Li et al., 2019; Mills et al., 2023).

In phase two, progressive unfreezing would be used to train local features on each hospital's local dataset at the hospital and either kept local or aggregated server-side. Our experiments are relatively lightweight and do not require large datacenter resources. Each hospital now runs the progressive-freezing schedule locally, validated in this study. Blocks in layer4 are unfrozen first, followed by layer3, while early blocks stay fixed. We synchronise only the unfrozen parameters via FedAvg. Approaches such as SmartFreeze (Wu et al., 2024) show that this approach is feasible and mitigates client drift.

The final phase could add a final local boosting by a site-specific fine tuning to account for center specific differences in acquisition parameters and calibration to the local scanners. Here, each site keeps the federated backbone frozen and refines a site-specific head or adapter (e.g. compare FedLP (Li et al., 2024)) that is evaluated through cross validation on the local dataset if dataset sizes allow.

To evaluate the overall model development, we could exchange the local models of the last phase in the end to evaluate the overall generalizability of the different local and global models on different datasets. This could help describing the overall learning capabilities but could also help identifying similarieties in different clinics.

Such an model approach could be implementated within many federated learing frameworks such as the FeatureCloud (Matschinske et al., 2023) or Flower (Beutel et al., 2022). The final model could be implemented in a local imaging workstation or a dashboard on a neonatology ward. In the future, the usage of ResNet-50 as basis could generally also allow for more enhanced mobile deployments on Android or WebAssembly. While such Web/Android-based solutions could enable the whole data processing on a mobile device to ensure that images remain entirely on the user's device and are never uploaded to external servers, they still face huge regulatory challenges for real world deployment.

Our results show that early BPD outcome prediction is achievable from routine admission radiographs using domain-specific pretraining and progressive fine-tuning. While single-center validation limits immediate generalizability, the approach's compatibility with small datasets, federated learning frameworks and open-source implementation enables broader applicability for resource-constrained medical imaging tasks.

## Data Availability

We will share our data with investigators whose proposed use of the data has been approved by an independent review committee (learned intermediary) identified for this purpose. Proposals should be directed to sascha.meyer@klinikum-karlsruhe.de. To gain access, data requestors will need to sign a data access agreement. Data will be available for 5 years.

## Code Availability

The PyTorch training and inference code, model weights, and configuration used for the ablation study is available on github.

## Abbreviations

AUROC area under receiver operating characteristic

IVH intraventricular hemorrhage

ROP retinopathy prematurorum

PVL periventricular leukomalacia

NEC necrotizing enterocolitis

BPD Bronchopulmonary dysplasia

ELBW extremely low birth weight

Sa02 oxygen saturation of arterial blood

CPAP continuous positive airway pressure

HFNC high flow nasal cannula

IRDS infant respiratory distress syndrome

AI artificial intelligence

CNNs convolutional neural networks

IVH intraventricular hemorrhage

NICU neonatal intensive care unit

TP true positive

FP false positive

TN true negative

FN false negative

**Acknowledgments**

The authors acknowledge HPC resources support with hardware funded by the DFG within project 469073465 and the Fox Foundation (MJFF-021418). In parts, this work was supported by the German Ministry of Research, Technology and Space, BMFTR (#01ZZ2005). The graphical abstract was created with biorender.com. OpenAI DALL·E 3 was used for the generation of the x-ray examples used in the graphical abstract.

**Conflict of Interest**

We declare no competing interests. S. Goedicke-Fritz and P. Flotho have been invited speakers of the company Chiesi to advice on Computer Vision for BPD prediction. Chiesi had no influence on study protocol, data compilation, and data analysis.

**Author Contributions**

Sybelle Goedicke-Fritz: Conceptualization; Project administration; Ethics approval; Data curation; Investigation; Writing - original draft; Writing - review & editing; Supervision (Clinical).

Michelle Bous: Conceptualization; Methodology; Ethics approval; Investigation; Data curation; Writing - original draft.

Annika Engel: Conceptualization; Data curation; Software; Writing - original draft.

Matthias Flotho: Visualization; Writing - review & editing; Formal analysis.

Pascal Hirsch: Writing - review & editing.

Hannah Wittig: Resources.

Dino Milanovic: Software; Data curation.

Dominik Mohr: Resources.

Mathias Kaspar: Writing – review & editing.

Sogand Nemat: Data curation; investigation.

Dorothea Kern: Data curation; investigation.

Arno Bücker: Resources.

Andreas Keller: Supervision (AI Analysis).

Sascha Meyer: Supervision (NeovitaA Study); Data curation; investigation; Funding acquisition (NeovitaA Study); Writing - review & editing.

Michael Zemlin: Supervision (Clinical); Data curation; Investigation; Conceptualization; Funding acquisition; Validation (Clinical); Writing - review & editing.

Philipp Flotho: Supervision (AI Analysis); Conceptualization; Formal analysis; Investigation; Data curation; Software; Writing - original draft; Writing - review & editing; Validation (AI Analysis)

## Tables

| **Experiment** | **AUROC** | **Accuracy** | **F1 Score** | **Sensitivity** | **Specificity** | **Precision** |
|---|---|---|---|---|---|---|
| XRV-ProgFreeze + LP + CutMix (40e) | **0.783** ± 0.095 | 0.686 ± 0.101 | 0.671 ± 0.111 | 0.654 ± 0.146 | 0.717 ± 0.128 | 0.706 ± 0.117 |
| XRV-ProgFreeze | 0.775 ± 0.097 | 0.688 ± 0.096 | 0.669 ± 0.105 | 0.640 ± 0.136 | 0.737 ± 0.143 | 0.723 ± 0.133 |
| XRV-ProgFreeze + CutMix | 0.771 ± 0.090 | **0.702** ± 0.082 | **0.687** ± 0.088 | **0.663** ± 0.128 | 0.741 ± 0.122 | 0.730 ± 0.110 |
| XRV-FullFT + LP + CutMix (40e) | 0.765 ± 0.090 | 0.697 ± 0.094 | 0.674 ± 0.104 | 0.636 ± 0.133 | 0.758 ± 0.119 | **0.731** ± 0.110 |
| XRV-ProgFreeze + LP | 0.762 ± 0.099 | 0.694 ± 0.097 | 0.671 ± 0.119 | 0.639 ± 0.150 | 0.749 ± 0.119 | 0.723 ± 0.115 |
| XRV-FullFT | 0.761 ± 0.094 | 0.690 ± 0.114 | 0.664 ± 0.135 | 0.630 ± 0.164 | 0.749 ± 0.149 | 0.725 ± 0.143 |
| XRV-FullFT + CutMix | 0.755 ± 0.094 | 0.693 ± 0.110 | 0.670 ± 0.124 | 0.641 ± 0.163 | 0.745 ± 0.143 | 0.726 ± 0.133 |
| XRV-ProgFreeze + LP + CutMix | 0.753 ± 0.096 | 0.684 ± 0.088 | 0.660 ± 0.102 | 0.625 ± 0.132 | 0.744 ± 0.119 | 0.717 ± 0.107 |
| RGB-FullFT | 0.752 ± 0.099 | 0.662 ± 0.087 | 0.607 ± 0.144 | 0.562 ± 0.197 | **0.761** ± 0.131 | 0.717 ± 0.123 |
| XRV-FullFT + LP | 0.751 ± 0.091 | 0.677 ± 0.099 | 0.651 ± 0.121 | 0.621 ± 0.159 | 0.732 ± 0.143 | 0.713 ± 0.133 |
| XRV-ProgFreeze + LP + CutMix (ProbeMix) | 0.748 ± 0.098 | 0.687 ± 0.090 | 0.664 ± 0.105 | 0.631 ± 0.134 | 0.744 ± 0.122 | 0.718 ± 0.108 |
| XRV-FullFT + LP + CutMix (ProbeMix) | 0.744 ± 0.096 | 0.672 ± 0.096 | 0.650 ± 0.110 | 0.618 ± 0.139 | 0.727 ± 0.138 | 0.705 ± 0.124 |
| XRV-FullFT + LP + CutMix | 0.743 ± 0.095 | 0.674 ± 0.092 | 0.648 ± 0.104 | 0.612 ± 0.139 | 0.736 ± 0.138 | 0.712 ± 0.124 |
| RGB-FullFT + CutMix | 0.724 ± 0.119 | 0.643 ± 0.088 | 0.597 ± 0.128 | 0.556 ± 0.177 | 0.731 ± 0.130 | 0.696 ± 0.141 |
| RGB-ProgFreeze | 0.717 ± 0.094 | 0.624 ± 0.086 | 0.557 ± 0.123 | 0.491 ± 0.153 | 0.757 ± 0.162 | 0.693 ± 0.151 |
| RGB-ProgFreeze + LP | 0.713 ± 0.113 | 0.632 ± 0.093 | 0.573 ± 0.149 | 0.528 ± 0.190 | 0.737 ± 0.182 | 0.695 ± 0.159 |
| RGB-ProgFreeze + CutMix | 0.712 ± 0.106 | 0.630 ± 0.098 | 0.539 ± 0.168 | 0.469 ± 0.193 | 0.791 ± 0.154 | 0.721 ± 0.176 |

*Table 1 Ablation study results for BPD prediction. Performance comparison of all experimental configurations sorted by AUROC. All values represent the mean ± standard deviation from 30 runs. Best values are put in bold.*

| Experiment | AUROC | Accuracy | F1 Score | Sensitivity | Specificity | Precision |
|---|---|---|---|---|---|---|
| XRV-ProgFreeze + LP + CutMix (40e) | **0.708** ± 0.363 | 0.790 ± 0.069 | 0.816 ± 0.080 | 0.819 ± 0.119 | 0.600 ± 0.321 | 0.831 ± 0.114 |
| XRV-ProgFreeze + CutMix | 0.705 ± 0.361 | 0.785 ± 0.074 | 0.813 ± 0.082 | 0.817 ± 0.096 | 0.591 ± 0.314 | 0.822 ± 0.121 |
| XRV-ProgFreeze | 0.705 ± 0.361 | 0.782 ± 0.066 | 0.808 ± 0.075 | 0.801 ± 0.089 | 0.600 ± 0.317 | 0.826 ± 0.114 |
| XRV-FullFT + LP + CutMix (40e) | 0.704 ± 0.361 | 0.795 ± 0.074 | 0.822 ± 0.083 | 0.830 ± 0.113 | 0.596 ± 0.321 | 0.829 ± 0.121 |
| XRV-FullFT + CutMix | 0.704 ± 0.361 | **0.802** ± 0.071 | **0.829** ± 0.076 | **0.839** ± 0.091 | **0.599** ± 0.319 | **0.831** ± 0.118 |
| XRV-FullFT | 0.703 ± 0.361 | 0.800 ± 0.069 | 0.827 ± 0.076 | 0.833 ± 0.082 | 0.599 ± 0.320 | **0.831** ± 0.119 |
| XRV-ProgFreeze + LP | 0.702 ± 0.360 | 0.796 ± 0.070 | 0.824 ± 0.073 | 0.831 ± 0.082 | 0.597 ± 0.317 | 0.829 ± 0.118 |
| XRV-ProgFreeze + LP + CutMix | 0.702 ± 0.360 | 0.793 ± 0.062 | 0.823 ± 0.070 | 0.831 ± 0.085 | 0.592 ± 0.317 | 0.827 ± 0.117 |
| XRV-FullFT + LP + CutMix | 0.702 ± 0.360 | 0.797 ± 0.063 | 0.828 ± 0.068 | 0.840 ± 0.090 | 0.594 ± 0.314 | 0.829 ± 0.114 |
| XRV-FullFT + LP + CutMix (ProbeMix) | 0.702 ± 0.360 | 0.798 ± 0.062 | 0.828 ± 0.070 | 0.840 ± 0.090 | 0.594 ± 0.314 | 0.828 ± 0.114 |
| XRV-ProgFreeze + LP + CutMix (ProbeMix) | 0.702 ± 0.360 | 0.792 ± 0.063 | 0.821 ± 0.072 | 0.831 ± 0.085 | 0.587 ± 0.313 | 0.823 ± 0.117 |
| XRV-FullFT + LP | 0.701 ± 0.360 | 0.799 ± 0.062 | 0.827 ± 0.070 | 0.837 ± 0.079 | 0.596 ± 0.316 | 0.829 ± 0.115 |
| RGB-ProgFreeze + CutMix | 0.693 ± 0.356 | 0.786 ± 0.067 | 0.823 ± 0.061 | 0.855 ± 0.111 | 0.554 ± 0.297 | 0.812 ± 0.111 |
| RGB-FullFT | 0.688 ± 0.352 | 0.793 ± 0.048 | 0.827 ± 0.059 | 0.858 ± 0.083 | 0.548 ± 0.299 | 0.810 ± 0.109 |
| RGB-FullFT + CutMix | 0.686 ± 0.354 | 0.769 ± 0.074 | 0.804 ± 0.075 | 0.815 ± 0.115 | 0.565 ± 0.303 | 0.811 ± 0.114 |
| RGB-ProgFreeze + LP | 0.682 ± 0.351 | 0.762 ± 0.088 | 0.790 ± 0.083 | 0.779 ± 0.128 | 0.593 ± 0.317 | 0.826 ± 0.114 |
| RGB-ProgFreeze | 0.676 ± 0.350 | 0.779 ± 0.078 | 0.811 ± 0.080 | 0.815 ± 0.108 | 0.578 ± 0.313 | 0.821 ± 0.116 |

*Table 2 Performance on predicting binned IRDS severity from radiographs. Results show model performance for classifying IRDS grades that were binarized (Grades I-II vs. III-IV based on majority vote of three experts). All values represent the mean ± standard deviation from 30 runs. Best values are put in bold and sorted by AUROC.*

## Figures

**Figure 1**: Cohort overview and study design. (**a**) Methodology: pretrained model selection (TorchXRayVision vs ImageNet), cohort analysis (163 ELBW infants), and systematic evaluation via 6 times repeated 5-fold cross-validation with ablation studies. (b-e) Population characteristics: birth weight and maternal age distributions, APGAR scores, and clinical characteristics prevalence.

**Figure 2**: Domain-specific pretraining improves BPD prediction. (**a**) XRV pretrained models (purple) outperform RGB/ImageNet models (blue) across all most metrics. Box plots show median, IQR, and 1.5 times IQR. (**b**) Test accuracy vs AUROC scatter plot: XRV models cluster in high-performance region and progressive freezing with linear probing and CutMix (40 epochs) achieves highest performance. (**c**) Confusion matrix components (TN: red, FP: orange, FN: purple, TP: teal) demonstrate balanced classification across approaches, with XRV models showing superior accuracy.

**Figure 3:** IRDS prediction shows different performance characteristics than BPD using our best model (XRV-ProgFreeze + LP + CutMix, 40 epochs). (**a**) Using the model that achieved highest AUROC for BPD (0.783), IRDS shows higher accuracy (0.790 vs 0.686) and F1 score (0.816 vs 0.671) but lower AUROC (0.708 vs 0.783). (**b**) Confusion matrices reveal IRDS achieves better sensitivity (81.9% vs 65.4%) at the cost of specificity (60.0% vs 71.7%). (**c**) Performance distributions across 510 runs. (**d**) Domain-specific pretraining improves both tasks. (**e,f**) IRDS models operate at high-sensitivity/low-specificity point. High sensitivity confirms severe IRDS features are visually distinct and learnable, yet these same features provide minimal prognostic value for BPD, demonstrating that BPD prediction requires different radiographic markers than IRDS grading.

## References

Ahmed, N., Kukleva, A., & Schiele, B. (2024). Orco: Towards better generalization via orthogonality and contrast for few-shot class-incremental learning. Proceedings of the IEEE/CVF Conference on Computer Vision and Pattern Recognition,

Ali, M. A., Maeda, R., Fujita, D., Miyahara, N., Namba, F., & Kobashi, S. (2025). Early prediction of bronchopulmonary dysplasia in preterm infants using chest X-rays through a comparative analysis of 13 CNN models across different post-birth days. *Discover Computing*, *28*(1), 20.

Askie, L. M., Henderson-Smart, D. J., Irwig, L., & Simpson, J. M. (2003). Oxygen-saturation targets and outcomes in extremely preterm infants. *New England Journal of Medicine*, *349*(10), 959-967.

Baraldi, E., & Filippone, M. (2007). Chronic lung disease after premature birth. *New England Journal of Medicine*, *357*(19), 1946-1955.

Barrington, K. J., Finer, N., & Pennaforte, T. (2017). Inhaled nitric oxide for respiratory failure in preterm infants. *Cochrane Database of Systematic Reviews*(1).

Baud, O., Laughon, M., & Lehert, P. (2021). Survival without bronchopulmonary dysplasia of extremely preterm infants: a predictive model at birth. *Neonatology*, *118*(4), 385-393.

Beutel, D. J., Topal, T., Mathur, A., Qiu, X., Fernandez-Marques, J., Gao, Y., Sani, L., Li, K. H., Parcollet, T., & de Gusmão, P. P. B. (2022). Flower: A friendly federated learning framework.

Bustos, A., Pertusa, A., Salinas, J.-M., & De La Iglesia-Vaya, M. (2020). Padchest: A large chest x-ray image dataset with multi-label annotated reports. *Medical image analysis*, *66*, 101797.

Chaves, J. M. Z., Huang, S.-C., Xu, Y., Xu, H., Usuyama, N., Zhang, S., Wang, F., Xie, Y., Khademi, M., & Yang, Z. (2024). Towards a clinically accessible radiology foundation model: open-access and lightweight, with automated evaluation. *arXiv preprint arXiv:2403.08002*.

Cho, H. W., Jung, S., Park, K. H., Choi, J. W., Heo, J. S., Kim, J., Yun, H., Yu, D., Son, J., & Choi, B. M. (2025). Deep-Learning-Based Multi-Class Classification for Neonatal Respiratory Diseases on Chest Radiographs in Neonatal Intensive Care Units. *Neonatology*.

Chou, H.-Y., Lin, Y.-C., Hsieh, S.-Y., Chou, H.-H., Lai, C.-S., Wang, B., & Tsai, Y.-S. (2024). Deep learning model for prediction of bronchopulmonary dysplasia in preterm infants using chest radiographs. *Journal of Imaging Informatics in Medicine*, *37*(5), 2063-2073.

Cohen, J. P., Hashir, M., Brooks, R., & Bertrand, H. (2020). On the limits of cross-domain generalization in automated X-ray prediction. Medical Imaging with Deep Learning,

Cohen, J. P., Viviano, J. D., Bertin, P., Morrison, P., Torabian, P., Guarrera, M., Lungren, M. P., Chaudhari, A., Brooks, R., & Hashir, M. (2022). TorchXRayVision: A library of chest X-ray datasets and models. International Conference on Medical Imaging with Deep Learning,

Darlow, B. A., Graham, P., & Rojas-Reyes, M. X. (2016). Vitamin A supplementation to prevent mortality and short-and long-term morbidity in very low birth weight infants. *Cochrane Database of Systematic Reviews*(8).

Demner-Fushman, D., Kohli, M. D., Rosenman, M. B., Shooshan, S. E., Rodriguez, L., Antani, S., Thoma, G. R., & McDonald, C. J. (2016). Preparing a collection of radiology examinations for distribution and retrieval. *Journal of the American Medical Informatics Association*, *23*(2), 304-310.

Dini, G., Ceccarelli, S., & Celi, F. (2024). Strategies for the prevention of bronchopulmonary dysplasia. *Frontiers in Pediatrics*, *12*, 1439265.

Doyle, L. W. (2021). Postnatal corticosteroids to prevent or treat bronchopulmonary dysplasia. *Neonatology*, *118*(2), 244-251.

Doyle, L. W., Ehrenkranz, R. A., & Halliday, H. L. (2014). Early (< 8 days) postnatal corticosteroids for preventing chronic lung disease in preterm infants. *Cochrane Database of Systematic Reviews*(5).

Flotho, P., Bhamborae, M. J., Grün, T., Trenado, C., Thinnes, D., Limbach, D., & Strauss, D. J. (2021). Multimodal data acquisition at SARS-CoV-2 drive through screening centers: Setup description and experiences in Saarland, Germany. *Journal of Biophotonics*, *14*(8), e202000512.

Flotho, P., Piening, M., Kukleva, A., & Steidl, G. (2025). T-FAKE: Synthesizing Thermal Images for Facial Landmarking. Proceedings of the Computer Vision and Pattern Recognition Conference,

Goedicke-Fritz, S., Härtel, C., Krasteva-Christ, G., Kopp, M. V., Meyer, S., & Zemlin, M. (2017). Preterm birth affects the risk of developing immune-mediated diseases. *Frontiers in immunology*, *8*, 1266.

Gortner, L., Misselwitz, B., Milligan, D., Zeitlin, J., Kollée, L., Boerch, K., Agostino, R., Van Reempts, P., Chabernaud, J.-L., & Bréart, G. (2011). Rates of bronchopulmonary dysplasia in very preterm neonates in Europe: results from the MOSAIC cohort. *Neonatology*, *99*(2), 112-117.

Han, Y.-S., Kim, S.-H., & Sung, T.-J. (2021). Impact of the definition of bronchopulmonary dysplasia on neurodevelopmental outcomes. *Scientific reports*, *11*(1), 22589.

He, K., Zhang, X., Ren, S., & Sun, J. (2016). Deep residual learning for image recognition. Proceedings of the IEEE conference on computer vision and pattern recognition,

Herting, E. (2013). Less invasive surfactant administration (LISA)—ways to deliver surfactant in spontaneously breathing infants. *Early human development*, *89*(11), 875-880.

Higgins, R. D., Jobe, A. H., Koso-Thomas, M., Bancalari, E., Viscardi, R. M., Hartert, T. V., Ryan, R. M., Kallapur, S. G., Steinhorn, R. H., & Konduri, G. G. (2018). Bronchopulmonary dysplasia: executive summary of a workshop. *The Journal of pediatrics*, *197*, 300-308.

Hwang, J. K., Kim, D. H., Na, J. Y., Son, J., Oh, Y. J., Jung, D., Kim, C.-R., Kim, T. H., & Park, H.-K. (2023). Two-stage learning-based prediction of bronchopulmonary dysplasia in very low birth weight infants: a nationwide cohort study. *Frontiers in Pediatrics*, *11*, 1155921.

Irvin, J., Rajpurkar, P., Ko, M., Yu, Y., Ciurea-Ilcus, S., Chute, C., Marklund, H., Haghgoo, B., Ball, R., & Shpanskaya, K. (2019). Chexpert: A large chest radiograph dataset with uncertainty labels and expert comparison. Proceedings of the AAAI conference on artificial intelligence,

Jensen, E. A., & Schmidt, B. (2014). Epidemiology of bronchopulmonary dysplasia. *Birth Defects Research Part A: Clinical and Molecular Teratology*, *100*(3), 145-157.

Jeon, G. W., Oh, M., & Chang, Y. S. (2021). Definitions of bronchopulmonary dysplasia and long-term outcomes of extremely preterm infants in Korean Neonatal Network. *Scientific reports*, *11*(1), 24349.

Jobe, A. H., & Bancalari, E. (2001). Bronchopulmonary dysplasia. *American journal of respiratory and critical care medicine*, *163*(7), 1723-1729.

Johnson, A. E., Pollard, T. J., Berkowitz, S. J., Greenbaum, N. R., Lungren, M. P., Deng, C.-y., Mark, R. G., & Horng, S. (2019). MIMIC-CXR, a de-identified publicly available database of chest radiographs with free-text reports. *Scientific data*, *6*(1), 317.

Joseph, S., & Joshi, H. (2024). ULMFiT: Universal Language Model Fine-Tuning for Text Classification. *International Journal of Advanced Medical Sciences and Technology (IJAMST)*, *4*.

Kanagaraj, U. K., Kulkarni, T., Kwan, E., Zhang, Q., Bone, J., & Shivananda, S. (2025). Validation of the NICHD Bronchopulmonary Dysplasia Outcome Estimator 2022 in a Quaternary Canadian NICU—A Single-Center Observational Study. *Journal of Clinical Medicine*, *14*(3), 696.

Ke, A., Ellsworth, W., Banerjee, O., Ng, A. Y., & Rajpurkar, P. (2021). CheXtransfer: performance and parameter efficiency of ImageNet models for chest X-Ray interpretation. Proceedings of the conference on health, inference, and learning,

Kornblith, S., Shlens, J., & Le, Q. V. (2019). Do better imagenet models transfer better? Proceedings of the IEEE/CVF conference on computer vision and pattern recognition,

Kukleva, A., Sener, F., Remelli, E., Tekin, B., Sauser, E., Schiele, B., & Ma, S. (2024). X-MIC: Cross-Modal Instance Conditioning for Egocentric Action Generalization. Proceedings of the IEEE/CVF Conference on Computer Vision and Pattern Recognition,

Kwok, T. C., Batey, N., Luu, K. L., Prayle, A., & Sharkey, D. (2023). Bronchopulmonary dysplasia prediction models: a systematic review and meta-analysis with validation. *Pediatr Res*, *94*(1), 43-54. https://doi.org/10.1038/s41390-022-02451-8

Laughon, M. M., Langer, J. C., Bose, C. L., Smith, P. B., Ambalavanan, N., Kennedy, K. A., Stoll, B. J., Buchter, S., Laptook, A. R., & Ehrenkranz, R. A. (2011). Prediction of bronchopulmonary dysplasia by postnatal age in extremely premature infants. *American journal of respiratory and critical care medicine*, *183*(12), 1715-1722.

Legate, G., Bernier, N., Page-Caccia, L., Oyallon, E., & Belilovsky, E. (2023). Guiding the last layer in federated learning with pre-trained models. *Advances in Neural Information Processing Systems*, *36*, 69832-69848.

Li, X., Huang, K., Yang, W., Wang, S., & Zhang, Z. (2019). On the convergence of fedavg on non-iid data. *arXiv preprint arXiv:1907.02189*.

Li, Y., Chen, H., Zhu, J., & Wang, Y. (2024). A Federated Learning Method Based on Linear Probing and Fine-Tuning. International Conference on Blockchain and Trustworthy Systems,

Lin, W., Kukleva, A., Sun, K., Possegger, H., Kuehne, H., & Bischof, H. (2022). Cycda: Unsupervised cycle domain adaptation to learn from image to video. European Conference on Computer Vision,

Ma, J., & Ye, H. (2016). Effects of permissive hypercapnia on pulmonary and neurodevelopmental sequelae in extremely low birth weight infants: a meta-analysis. *Springerplus*, *5*(1), 764.

Matschinske, J., Späth, J., Bakhtiari, M., Probul, N., Kazemi Majdabadi, M. M., Nasirigerdeh, R., Torkzadehmahani, R., Hartebrodt, A., Orban, B.-A., & Fejér, S.-J. (2023). The FeatureCloud platform for federated learning in biomedicine: unified approach. *Journal of Medical Internet Research*, *25*, e42621.

McMahan, B., Moore, E., Ramage, D., Hampson, S., & y Arcas, B. A. (2017). Communication-efficient learning of deep networks from decentralized data. Artificial intelligence and statistics,

Meyer, S., Bay, J., Franz, A. R., Ehrhardt, H., Klein, L., Petzinger, J., Binder, C., Kirschenhofer, S., Stein, A., & Hüning, B. (2024). Early postnatal high-dose fat-soluble enteral vitamin A supplementation for moderate or severe bronchopulmonary dysplasia or death in extremely low birthweight infants (NeoVitaA): A multicentre, randomised, parallel-group, double-blind, placebo-controlled, investigator-initiated phase 3 trial. *The Lancet Respiratory Medicine*, *12*(7), 544-555.

Meyer, S., & Gortner, L. (2017). Up-date on the NeoVitaA Trial: Obstacles, challenges, perspectives, and local experiences. *Wiener medizinische Wochenschrift*, *167*, 264-270.

Meyer, S., Gortner, L., & Investigators, N. T. (2014). Early postnatal additional high-dose oral vitamin A supplementation versus placebo for 28 days for preventing bronchopulmonary dysplasia or death in extremely low birth weight infants. *Neonatology*, *105*(3), 182-188.

Mills, J., Hu, J., & Min, G. (2023). Faster federated learning with decaying number of local SGD steps. *IEEE Transactions on Parallel and Distributed Systems*, *34*(7), 2198-2207.

Nguyen, H. Q., Lam, K., Le, L. T., Pham, H. H., Tran, D. Q., Nguyen, D. B., Le, D. D., Pham, C. M., Tong, H. T., & Dinh, D. H. (2022). VinDr-CXR: An open dataset of chest X-rays with radiologist's annotations. *Scientific data*, *9*(1), 429.

Nuytten, A., Behal, H., Duhamel, A., Jarreau, P.-H., Torchin, H., Milligan, D., Maier, R. F., Zemlin, M., Zeitlin, J., & Truffert, P. (2020). Postnatal corticosteroids policy for very preterm infants and bronchopulmonary dysplasia. *Neonatology*, *117*(3), 308-315.

Polin, R. A., Carlo, W. A., Fetus, C. o., Newborn, Papile, L.-A., Polin, R. A., Carlo, W., Tan, R., Kumar, P., Benitz, W., & Eichenwald, E. (2014). Surfactant replacement therapy for preterm and term neonates with respiratory distress. *Pediatrics*, *133*(1), 156-163.

Prodanovic, T., Petrovic Savic, S., Prodanovic, N., Simovic, A., Zivojinovic, S., Djordjevic, J. C., & Savic, D. (2024). Advanced diagnostics of respiratory distress syndrome in premature infants treated with surfactant and budesonide through computer-assisted chest x-ray analysis. *Diagnostics*, *14*(2), 214.

Romijn, M., Dhiman, P., Finken, M. J., van Kaam, A. H., Katz, T. A., Rotteveel, J., Schuit, E., Collins, G. S., Onland, W., & Torchin, H. (2023). Prediction models for bronchopulmonary dysplasia in preterm infants: a systematic review and meta-analysis. *The Journal of pediatrics*, *258*, 113370.

Russakovsky, O., Deng, J., Su, H., Krause, J., Satheesh, S., Ma, S., Huang, Z., Karpathy, A., Khosla, A., & Bernstein, M. (2015). Imagenet large scale visual recognition challenge. *International journal of computer vision*, *115*, 211-252.

Sardesai, S., Biniwale, M., Wertheimer, F., Garingo, A., & Ramanathan, R. (2017). Evolution of surfactant therapy for respiratory distress syndrome: past, present, and future. *Pediatric Research*, *81*(1), 240-248.

Schmidt, B., Asztalos, E. V., Roberts, R. S., Robertson, C. M., Sauve, R. S., Whitfield, M. F., & Investigators, T. o. I. P. i. P. (2003). Impact of bronchopulmonary dysplasia, brain injury, and severe retinopathy on the outcome of extremely low-birth-weight infants at 18 months: results from the trial of indomethacin prophylaxis in preterms. *Jama*, *289*(9), 1124-1129.

Schmidt, B., Roberts, R. S., Davis, P., Doyle, L. W., Barrington, K. J., Ohlsson, A., Solimano, A., & Tin, W. (2006). Caffeine therapy for apnea of prematurity. *New England Journal of Medicine*, *354*(20), 2112-2121.

Shih, G., Wu, C. C., Halabi, S. S., Kohli, M. D., Prevedello, L. M., Cook, T. S., Sharma, A., Amorosa, J. K., Arteaga, V., & Galperin-Aizenberg, M. (2019). Augmenting the national institutes of health chest radiograph dataset with expert annotations of possible pneumonia. *Radiology: Artificial Intelligence*, *1*(1), e180041.

Shvetsova, N., Kukleva, A., Schiele, B., & Kuehne, H. (2023). In-style: Bridging text and uncurated videos with style transfer for text-video retrieval. Proceedings of the IEEE/CVF International Conference on Computer Vision,

Slutsky, A. S. (1999). Lung injury caused by mechanical ventilation. *Chest*, *116*, 9S-15S.

Smith, V. C., Zupancic, J. A., McCormick, M. C., Croen, L. A., Greene, J., Escobar, G. J., & Richardson, D. K. (2005). Trends in severe bronchopulmonary dysplasia rates between 1994 and 2002. *The Journal of pediatrics*, *146*(4), 469-473.

Srivatsa, B., Srivatsa, K. R., & Clark, R. H. (2023). Assessment of validity and utility of a bronchopulmonary dysplasia outcome estimator. *Pediatric Pulmonology*, *58*(3), 788-793.

Stichtenoth, G., Demmert, M., Bohnhorst, B., Stein, A., Ehlers, S., Heitmann, F., Rieger-Fackeldey, E., Olbertz, D., Roll, C., & Emeis, M. (2012). Major contributors to hospital mortality in very-low-birth-weight infants: data of the birth year 2010 cohort of the German Neonatal Network. *Klinische Pädiatrie*, 276-281.

Sweet, D. G., Carnielli, V. P., Greisen, G., Hallman, M., Klebermass-Schrehof, K., Ozek, E., Te Pas, A., Plavka, R., Roehr, C. C., & Saugstad, O. D. (2023). European consensus guidelines on the management of respiratory distress syndrome: 2022 update. *Neonatology*, *120*(1), 3-23.

Thomas, W., & Speer, C. P. (2005). Management of infants with bronchopulmonary dysplasia in Germany. *Early human development*, *81*(2), 155-163.

Tyson, J. E., Wright, L. L., Oh, W., Kennedy, K. A., Mele, L., Ehrenkranz, R. A., Stoll, B. J., Lemons, J. A., Stevenson, D. K., & Bauer, C. R. (1999). Vitamin A supplementation for extremely-low-birth-weight infants. *New England Journal of Medicine*, *340*(25), 1962-1968.

Verder, H., Heiring, C., Ramanathan, R., Scoutaris, N., Verder, P., Jessen, T. E., Höskuldsson, A., Bender, L., Dahl, M., & Eschen, C. (2021). Bronchopulmonary dysplasia predicted at birth by artificial intelligence. *Acta Paediatrica*, *110*(2), 503-509.

Wang, X., Peng, Y., Lu, L., Lu, Z., Bagheri, M., & Summers, R. M. (2017). Chestx-ray8: Hospital-scale chest x-ray database and benchmarks on weakly-supervised classification and localization of common thorax diseases. Proceedings of the IEEE conference on computer vision and pattern recognition,

Watterberg, K. L., Gerdes, J. S., Cole, C. H., Aucott, S. W., Thilo, E. H., Mammel, M. C., Couser, R. J., Garland, J. S., Rozycki, H. J., & Leach, C. L. (2004). Prophylaxis of early adrenal insufficiency to prevent bronchopulmonary dysplasia: a multicenter trial. *Pediatrics*, *114*(6), 1649-1657.

Wu, Y., Li, L., Tian, C., Chang, T., Lin, C., Wang, C., & Xu, C.-Z. (2024). Heterogeneity-aware memory efficient federated learning via progressive layer freezing. 2024 IEEE/ACM 32nd International Symposium on Quality of Service (IWQoS),

Xing, W., He, W., Li, X., Chen, J., Cao, Y., Zhou, W., Shen, Q., Zhang, X., & Ta, D. (2022). Early severity prediction of BPD for premature infants from chest X-ray images using deep learning: A study at the 28th day of oxygen inhalation. *Computer Methods and Programs in Biomedicine*, *221*, 106869.

Yeh, T. F., Lin, Y. J., Lin, H. C., Huang, C. C., Hsieh, W. S., Lin, C. H., & Tsai, C. H. (2004). Outcomes at school age after postnatal dexamethasone therapy for lung disease of prematurity. *New England Journal of Medicine*, *350*(13), 1304-1313.

Yun, S., Han, D., Oh, S. J., Chun, S., Choe, J., & Yoo, Y. (2019). Cutmix: Regularization strategy to train strong classifiers with localizable features. Proceedings of the IEEE/CVF international conference on computer vision,

Zawacki, A., Wu, C., Shih, G., Elliott, J., Fomitchev, M., Hussain, M., Lakhani, P., Culliton, P., & Bao, S. (2019). Siim-acr pneumothorax segmentation. *Mohannad ParasLakhani Hussain*.

Zea-Vera, A., & Ochoa, T. J. (2015). Challenges in the diagnosis and management of neonatal sepsis. *Journal of tropical pediatrics*, *61*(1), 1-13.

Zemlin, M., Berger, A., Franz, A., Gille, C., Härtel, C., Küster, H., Müller, A., Pohlandt, F., Simon, A., & Merz, W. (2019). Bakterielle Infektionen bei Neugeborenen. Leitlinie der GNPI, DGPI, DGKJ und DGGG.(S2k-Level, AWMF-Leitlinien-Register-Nr. 024/008, April 2018). *Zeitschrift für Geburtshilfe und Neonatologie*, *223*(03), 130-144.

Figure 1 – Model Flow / Ablation

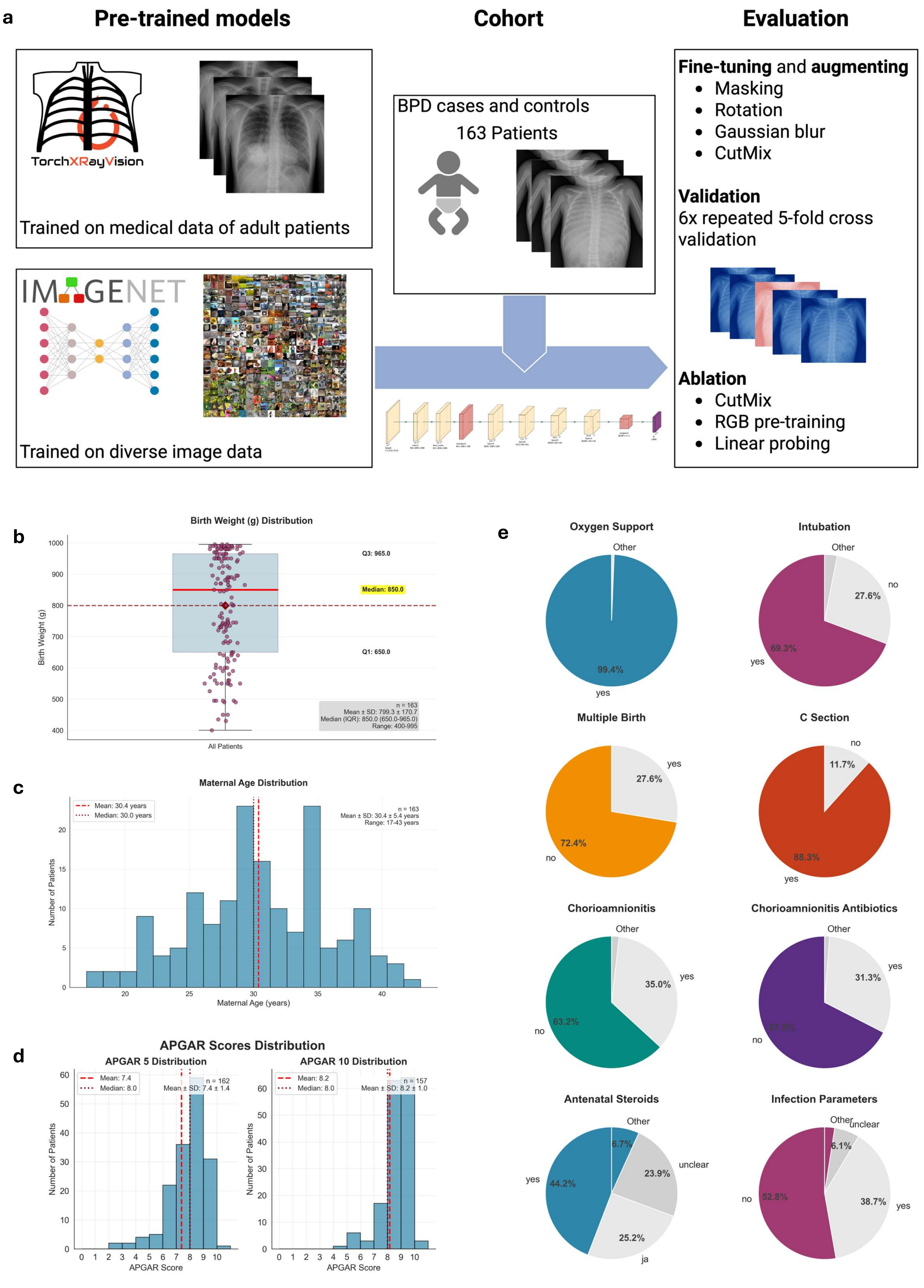

Figure 2 – PBD prediction ablation study

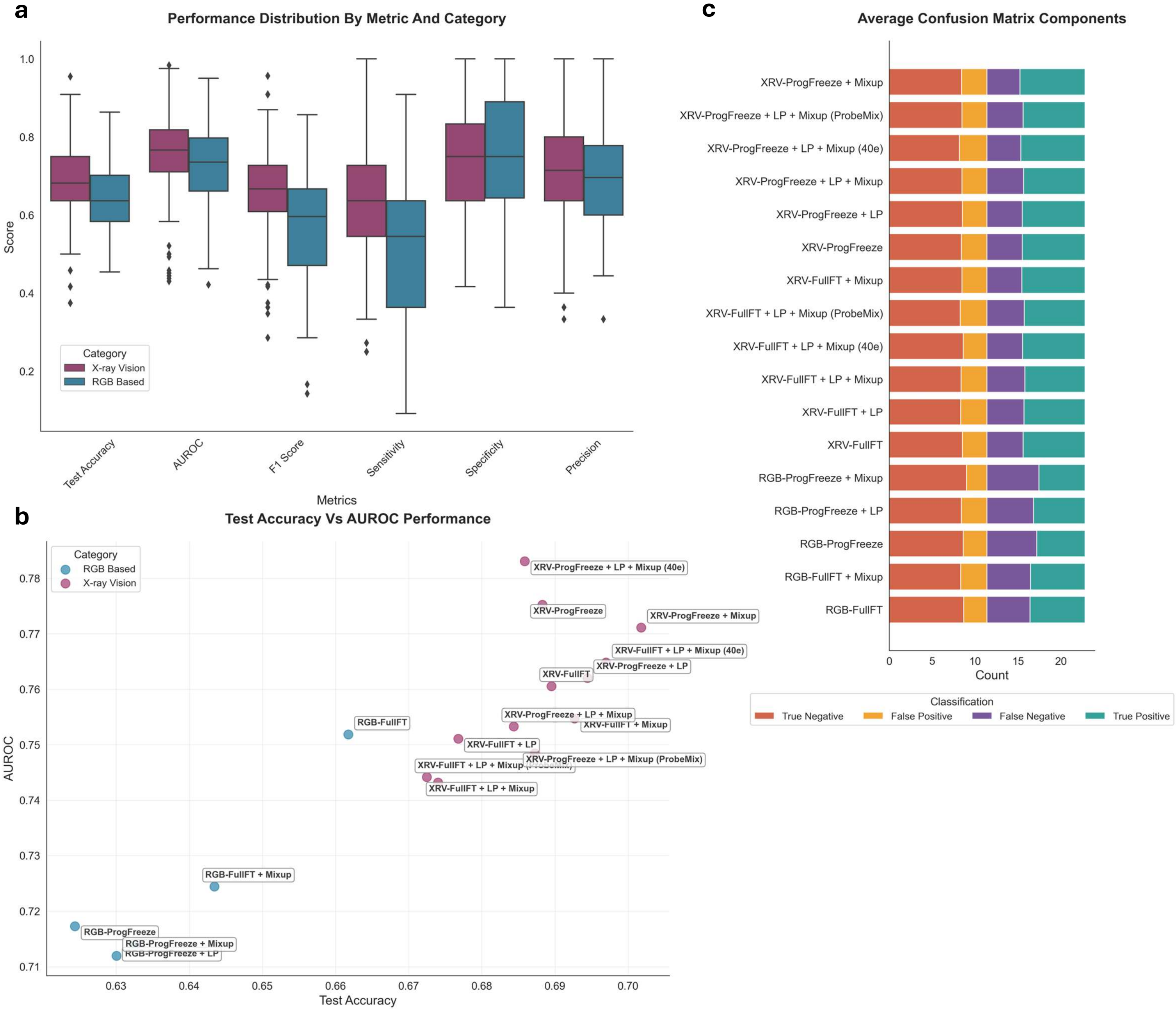

Figure 3 – PBD vs. IRDS: comprehensive Comparison analysis

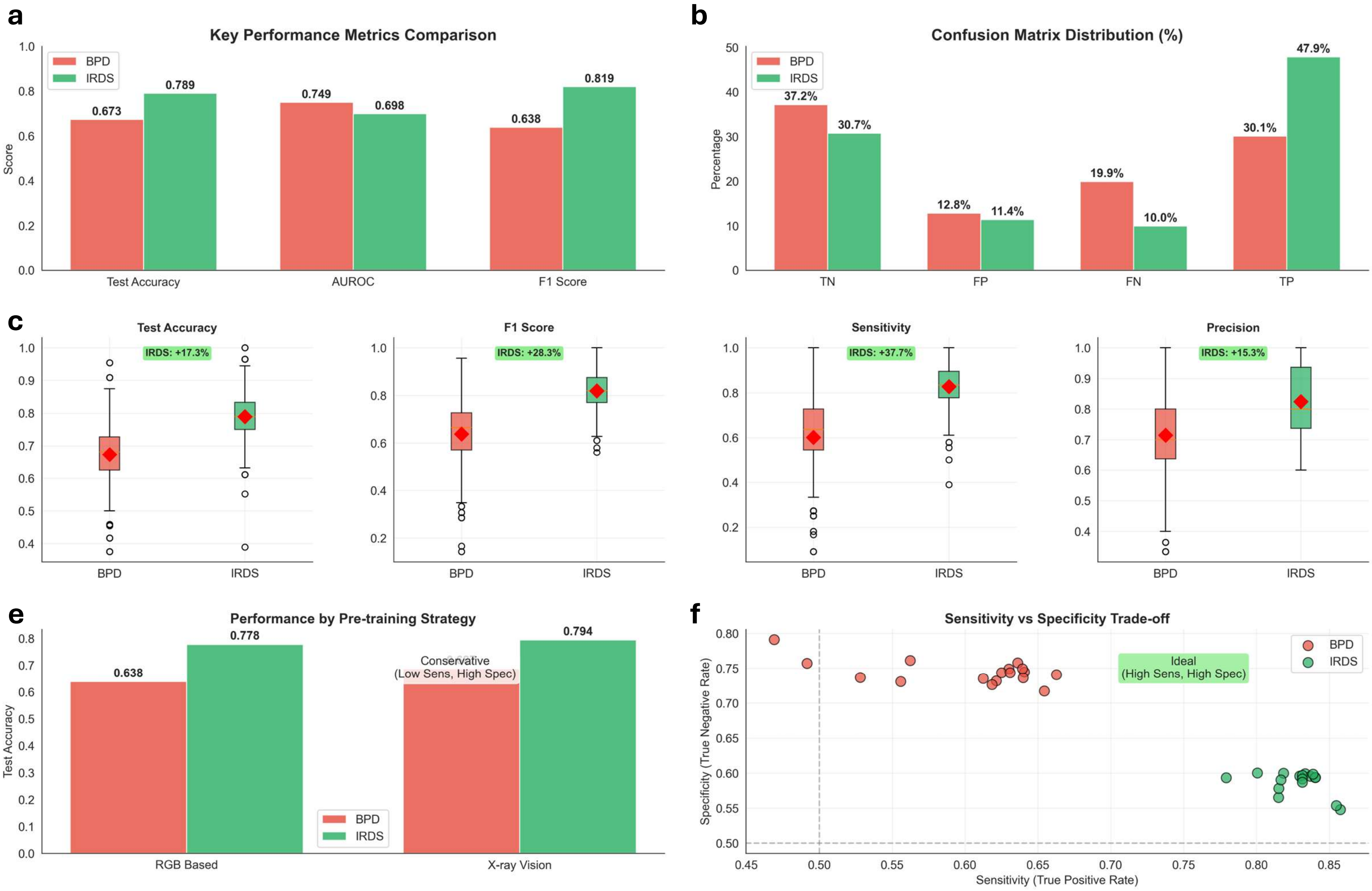